\documentclass{article}

\usepackage{microtype}
\usepackage{graphicx}
\usepackage{booktabs} 
\usepackage{hyperref}

\usepackage{amssymb}
\usepackage{mathtools}
\usepackage{amsthm}
\usepackage{algpseudocode}
\usepackage{color-edits}

\usepackage{enumitem}
\usepackage{enumitem}
\usepackage{subcaption}

\usepackage[accepted]{icml2024}

\theoremstyle{plain}

\theoremstyle{definition}

\theoremstyle{remark}

\usepackage{amsmath,amsfonts,bm}

\def\1{\bm{1}}

\DeclareMathAlphabet{\mathsfit}{\encodingdefault}{\sfdefault}{m}{sl}
\SetMathAlphabet{\mathsfit}{bold}{\encodingdefault}{\sfdefault}{bx}{n}

\DeclareMathOperator*{\argmax}{arg\,max}

\addauthor{natalie}{red}
\addauthor{jerry}{blue}
\addauthor{max}{magenta}
\addauthor{eytan}{purple}

\newcommand{\ex}{\textcolor{red}{\mathcal{E_X}}}
\newcommand{\ey}{\textcolor{blue}{\mathcal{E_Y}}}
\newcommand{\hhat}{\textcolor{red}{\hat{h}}}
\newcommand{\ghat}{\textcolor{blue}{\hat{g}}}
\newcommand{\ftrue}{\textcolor[HTML]{800080}{f}}
\newcommand{\htrue}{\textcolor{red}{h}}
\newcommand{\gtrue}{\textcolor{blue}{g}}

\usepackage[textsize=tiny]{todonotes}

\icmltitlerunning{Joint Composite Latent Space Bayesian Optimization}

\begin{document}

\twocolumn[
\icmltitle{Joint Composite Latent Space Bayesian Optimization}

\icmlsetsymbol{equal}{*}

\begin{icmlauthorlist}
\icmlauthor{Natalie Maus}{penn}
\icmlauthor{Zhiyuan Jerry Lin}{meta}
\icmlauthor{Maximilian Balandat}{meta}
\icmlauthor{Eytan Bakshy}{meta}
\end{icmlauthorlist}

\icmlaffiliation{penn}{Department of Computer and Information Science, University of Pennsylvania}
\icmlaffiliation{meta}{Meta}

\icmlcorrespondingauthor{Natalie Maus}{nmaus@seas.upenn.edu}

\icmlkeywords{Machine Learning, Bayesian optimization, composite function optimization}

\vskip 0.3in
]

\printAffiliationsAndNotice{} 
\bibliographystyle{icml2024}

\begin{abstract}
Bayesian Optimization (BO) is a technique for sample-efficient black-box optimization that employs probabilistic models to identify promising inputs 
for evaluation. When dealing with composite-structured functions such as $f=g \circ h$, evaluating a specific location $x$ yields observations of both the final outcome $f(x) = g(h(x))$ as well as the intermediate output(s) $h(x)$. Previous research has shown that integrating information from these intermediate outputs can enhance BO performance substantially. However, existing methods struggle if the outputs $h(x)$ are high-dimensional. Many relevant problems fall into this setting, including in the context of generative AI, molecular design, or robotics. 
To effectively tackle these challenges, we introduce Joint Composite Latent Space Bayesian Optimization (JoCo), a novel framework that jointly trains neural network encoders and probabilistic models to adaptively compress high-dimensional input and output spaces into manageable latent representations. This enables effective BO on these compressed representations, allowing JoCo to outperform other state-of-the-art methods in high-dimensional BO on a wide variety of simulated and real-world problems.
\end{abstract}

\section{Introduction}
\label{sec:introduction}

Many problems in engineering and science involve optimizing expensive-to-evaluate black-box functions. Bayesian Optimization (BO) has emerged as a sample-efficient approach to tackling this challenge.  
At a high level, BO builds a probabilistic \emph{surrogate model}, often a Gaussian Process, of the unknown function based on observed evaluations and then recommends the next query point(s) by optimizing an \emph{acquisition function} that leverages probabilistic model predictions to guide the exploration-exploitation tradeoff.
While the standard black-box approach is effective across many domains~\citep{frazier2016bayesian, packwood2017bayesian, zhang2020bayesian, calandra2016bayesian, letham2019constrained, mao2020real}, it does not make use of rich data that may be available when objectives may be stated in terms of a composite function 
$f = g \circ h$. In this setting, not only the final objective $f(x) =g(h(x))$, but also the outputs of the intermediate function, $h(x)$, can be observed upon evaluation, providing additional information that can be exploited for optimization.

While recent scientific advances~\citep{astudillo2019composite,lin2022preference} attempt to take advantage of this structure, they falter when $h$ maps to a high-dimensional intermediate outcome space, a common occurrence in a variety of applications.
For example, when optimizing foundational ML models  
with text prompts as inputs, intermediate outputs may be complex data types such as images or text and the objective may be to generate images of texts of a specific style.
In aerodynamic design problems, a high-dimensional input space of geometry and flow conditions are optimized to achieve specific objectives, e.g., minimizing drag while maintaining lift, defined over a high-dimensional output space of pressure and velocity fields~\citep{zawawi2018review, lomax2002fundamentals}.

Intuitively, the wealth of information contained in such high-dimensional intermediate data should pave the way for more efficient resolution of the task at hand.
However, to our knowledge, little literature exists on leveraging this potential efficiency gain when optimizing functions with high-dimensional intermediate outputs over high-dimensional input spaces. 
To close this gap, we introduce \textit{JoCo}, a new algorithm for \textit{\textbf{Jo}int \textbf{Co}mposite Latent Space Bayesian Optimization}.
Unlike standard BO, which constructs a surrogate model only for the full mapping $f$, \text{JoCo} simultaneously trains probabilistic models both for capturing the behavior of the black-box function and for compressing the high-dimensional intermediate output space. In doing so, it effectively leverages this additional information, yielding a method that substantially outperforms existing high-dimensional BO algorithms on problems with composite structure. 

\newpage
Our main contributions are:
\begin{enumerate}
    \item We introduce JoCo, a new algorithm for composite BO with high-dimensional input and output spaces. To our knowledge, JoCo is the first composite BO method capable of scaling to problems with very high-dimensional intermediate outputs.  
    \item We demonstrate that JoCo significantly outperforms other state-of-the-art baselines on a number of synthetic and real-world problems.
    \item We leverage JoCo to effectively perform black-box adversarial attacks on generative text and image models,  challenging settings with input and intermediate output dimensions in the thousands and hundreds of thousands, respectively. 
\end{enumerate}

\section{High-Dimensional Composite Objective Optimization}
\label{sec:hdcoo}

We consider the optimization of a \emph{composite} objective function
$\ftrue: \mathcal{X} \rightarrow \mathbb{R}$ defined as $\ftrue = \gtrue \circ \htrue$
where $\htrue: \mathcal{X} \rightarrow \mathcal{Y}$ and $\gtrue: \mathcal{Y} \rightarrow \mathbb{R}$. 
At least one of~$\htrue$ and~$\gtrue$ is expensive to evaluate, making it challenging to apply classic numerical optimization algorithms 
that generally require a large number of function evaluations. 
The key complication compared to more conventional composite BO settings is that inputs and intermediate outputs reside in high-dimensional vector spaces. Namely, $\mathcal{X} \subset \mathbb{R}^d$ and $\mathcal{Y} \subset \mathbb{R}^m$ for some large~$d$ and~$m$.
Concretely, the optimization problem we aim to solve is to identify $\mathbf{x}^* \in \mathcal{X}$ such that 
\begin{equation}
    \label{eq:bocf_formulation}
    \mathbf{x}^* \in \argmax_{\mathbf{x} \in \mathcal{X}} \ftrue(\mathbf{x}) = \argmax_{\mathbf{x} \in \mathcal{X}} 
 \gtrue(\htrue(\mathbf{x})).
\end{equation}

For instance, consider the scenario of optimizing generative AI models where $\mathcal{X}$ represents all possible text prompts of some maximum length (e.g., via vector embeddings for string sequences). 
The function $\htrue: \mathcal{X} \rightarrow \mathcal{Y}$ could map these text prompts to generated images, and the objective, represented by $\gtrue: \mathcal{Y} \rightarrow \mathbb{R}$, quantifies the probability of the generated image containing specific content (e.g., a dog).

Combining composite function optimization and high-dimensional BO inherits challenges from both domains, exacerbating some of them. 
The primary difficulty with high-dimensional~$\mathcal{X}$ and~$\mathcal{Y}$ is that the Gaussian Process (GP) models typically employed in BO do not perform well in this setting due to all observations being ``far away'' from each other~\citep{jiang2022gaussian, djolonga2013high}. 
In addition, in higher dimensions, identifying the correct kernel and hyperparameters becomes more difficult. 
When dealing with complex data structures such as texts or images, explicitly specifying the appropriate kernel might be even more challenging.
Furthermore, while BO typically assumes a known search space (often a hypercube), the structure and manifold of the intermediate space $\mathcal{Y}$ is generally unknown, complicating the task of accommodating high-dimensional modeling and optimization.

\subsection{Related Work}
\label{subsec:hdcoo:related_work}

\paragraph{Bayesian Optimization of Composite Functions}
\citet{astudillo2019composite} pioneered this area by proposing a method that exploits composite structure in objectives to improve sample efficiency.
This work is a specific instance of grey-box BO, which extends the classical BO setup to treat the objective function as partially observable and modifiable~\citep{astudillo2021thinking}. 
Grey-box BO methods, particularly those focusing on composite functions, have shown dramatic performance gains by exploiting known structure in the objective function.

For example,~\citet{astudillo2021bayesian} propose a framework for optimizing not just a composite function, but a much more complex, interdependent network of functions. \citet{maddox2021hdbo} tackled the issue of high-dimensional outputs in composite function optimization. They proposed a technique that exploits Kronecker structure in the covariance matrices when using Matheron's identity to optimize composite functions with tens of thousands of correlated outputs. However, scalability in the number of observations is limited (to the hundreds) due to high computational and memory requirements. 

\citet{candelieri2023wasserstein} propose to map the original problem into a space of discrete probability distributions measured with a Wasserstein metric, and by doing so show performance gains compared to traditional approaches, especially as the search space dimension increases.  
In the context of incorporating qualitative human feedback,~\citet{lin2022preference} introduced Bayesian Optimization with Preference Exploration (BOPE), which use preference learning leveraging pairwise comparisons between outcome vectors to reducing both experimental costs and time. 
This approach is especially useful when the function $g$ is not directly evaluable but can be elicited from human decision makers. 

While the majority of existing research on BO of composite structures focuses on leveraging pre-existing knowledge of objective structures, advancements in representation learning methods, such as deep kernel learning~\citep{wilson2016deep, wilson2016stochastic}, offer a new avenue. These methods enable the creation of learned latent representations for GP models. 
Despite this potential, there has been limited effort to explicitly utilize these expressive latent structures to enhance and scale up grey-box optimization.

\paragraph{Bayesian Optimization over High-Dimensional Input Spaces}
Optimizing black-box functions over high-dimensional domains $\mathcal{X}$ poses a unique set of challenges. Conventional BO strategies struggle with optimization tasks in spaces exceeding 15-20 continuous dimensions~\citep{wang2016rembo}. 
Various techniques have been developed to scale BO to higher dimensions, including but not limited to approaches that exploit low-dimensional additive structures~\citep{2015_Kandasamy, 2017_Gardner}, variable selection~\citep{eriksson2021saasbo, 2022_Song}, and trust region optimization~\citep{eriksson2019turbo}. 
Random embeddings were initially proposed as a solution for high-dimensional BO by~\citet{wang2016rembo} and expanded upon in later works (e.g., \citet{2017_Rana, nayebi19hesbo, letham2020re, 2020_Binois, 2022_Papenmeier}. 

Leveraging nonlinear embeddings based on autoencoders, \citet{2018_Gomez} spurred substantial research activity. Subsequent works have extended this ``latent space BO'' framework to incorporate label supervision and constraints on the latent space~\citep{2020_Griffiths, moriconi2020high, 2021_Notin, 2013_Snoek, 2019_Zhang, 2018_Eissman, 2020_Tripp, 2021_Siivola, 2020_Chen, 2021_Grosnit, 2022_Stanton, 2022_Maus, 2023_Maus, 2023_Yin}. However, these approaches are limited in that they require a large corpus of initial unlabeled data to pre-train the autoencoder.

\section{Method}
\label{sec:method}

\subsection{Intuition}
\label{subsec:method:intuition}

One may choose to directly apply standard high-dimensional Bayesian optimization methods such as TuRBO~\citep{eriksson2019turbo} or SAASBO~\citep{eriksson2021saasbo} to the problem~\eqref{eq:bocf_formulation}, ignoring the fact that $\ftrue$ has a composite structure and discarding the intermediate information~$\htrue(x)$.
To take advantage of composite structure,~\citet{astudillo2019composite} suggest to model $\htrue$ and $\gtrue$ separately. However, a high-dimensional space $\mathcal{Y}$ poses significant computational challenges for their and other existing methods.

To tackle this problem, we can follow the latent space BO literature to map the original high-dimensional intermediate output space $\mathcal{Y}$ into a low-dimensional manifold $\hat{\mathcal{Y}}$ such that modeling and optimization becomes feasible on $\hat{\mathcal{Y}}$.  
Common choices of such mappings include principal component analysis and variational autoencoders.
One key issue with these latent space methods is that they require an accurate latent representation for the \emph{original} space. 
This is a fundamental limitation that prevents us from further compressing the latent space into an even lower-dimensional space without losing too much information. 

In the context of composite BO, reconstructing the intermediate output is not actually a goal but merely a means to an end.
Instead, our actual goal is to map the intermediate output to a low-dimensional embedding that retains information relevant to the optimization goal, namely the final function value $\ftrue(\mathbf{x})$, and but not necessarily information unrelated to the optimization target.

By using the function value as supervisory information, we are able to learn, refine, and optimize both the probabilistic surrogate models and latent space encoders \emph{jointly} and \emph{continuously} as the optimization proceeds. 

\subsection{Joint Composite Latent Space Bayesian Optimization (JoCo)}
\label{subsec:method:joco}

\begin{figure}[t]
    \centering
    \includegraphics[width=\columnwidth]{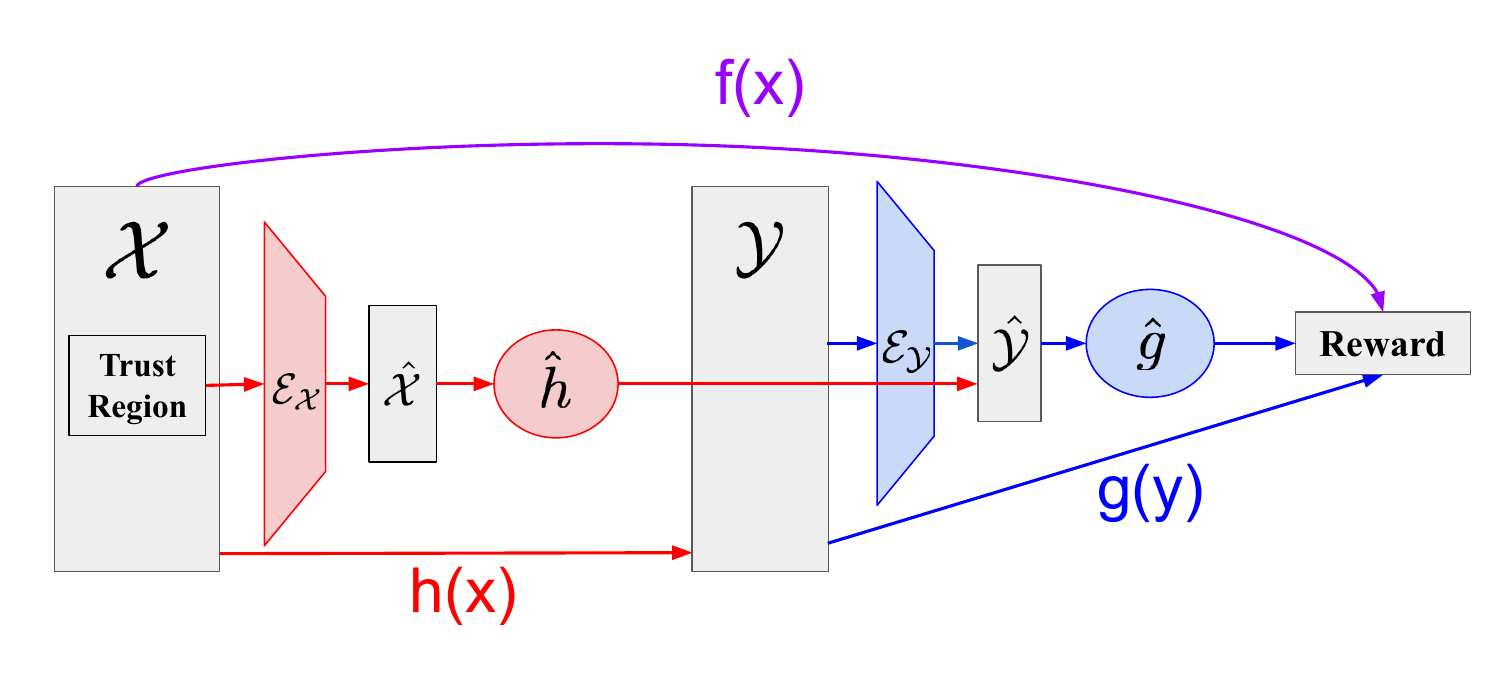}
    \caption{JoCo architecture:
Two NN encoders, $\ex$ and $\ey$, embed the high-dimensional input and intermediate output spaces into lower-dimensional latent spaces, $\hat{\mathcal{X}}$ and $\hat{\mathcal{Y}}$, respectively.
The latent probabilistic model $\hhat$ maps the embedded input space to a distribution over the embedded intermediate output space $\hat{\mathcal{Y}}$, while~$\ghat$ maps $\hat{\mathcal{Y}}$ to a distribution over possible composite function values.
Together, these components enable effective high-dimensional optimization by jointly learning representations that enable accurate prediction and optimization of the composite function~$\ftrue$.}
    \label{fig:joco_architecture}
\end{figure}

Figure~\ref{fig:joco_architecture} illustrates JoCo's architecture and Algorithm~\ref{alg:joco} outlines JoCo's procedures.  
\begin{algorithm*}
\caption{JoCo}
\label{alg:joco}
\begin{algorithmic}
\Require Input space $\mathcal{X}$, Number of TS samples $N_{\text{sample}}$, Initial data size $n$, number of iterations $N$ \State \textbf{Generate Initial Data:} $\mathcal{D} = \{(\mathbf{x}_1, \mathbf{y}_1, f(\mathbf{x}_1)), \dots, (\mathbf{x}_{n}, \mathbf{y}_{n}, f(\mathbf{x}_{n}))\}$ with $n$ random points.
\State \textbf{Fit Initial Models:} Initialize $\ex$, $\ey$, $\hhat$, $\ghat$ on $\mathcal{D}$ by minimizing ~\eqref{eq:joco_loss}.
\State \textbf{JoCo Optimization Loop:} 
\For{$i = 1, 2, \ldots, N$}
    \State $\mathbf{x}_{i} \gets \textsc{TS}(SS=\text{TuRBO Trust Region}, N_{\text{sample}}, \ex, \hhat, \ghat)$
    \State Evaluate $\mathbf{x}_{i}$ and observe $\mathbf{y}_{i}$ and $f(\mathbf{x}_{i})$.
    \State $\mathcal{D} \gets \mathcal{D} \cup \left\{(\mathbf{x}_{i}, \mathbf{y}_{i}, f(\mathbf{x}_{i}) \right\}$
    \State Update $\ex$, $\ey$, $\hhat$, and $\ghat$ jointly using the latest $N_b$
    data points by minimizing ~\eqref{eq:joco_loss} on $\mathcal{D}$. 
\EndFor
\State Find $\mathbf{x}_{\text{best}}$ such that $f(\mathbf{x}_{\text{best}})$ is the maximum in $\mathcal{D}$
\State \Return $\mathbf{x}_{\text{best}}$
\end{algorithmic}
\end{algorithm*}
Unlike conventional BO with a single probabilistic surrogate model, JoCo consists of four core components:
\begin{enumerate}[leftmargin=18pt, labelwidth=!, labelindent=0pt]
    \item \textbf{Input NN encoder $\ex: \mathcal{X} \to \hat{\mathcal{X}}$.} $\ex$ projects the input space $\mathbf{x} \in \mathcal{X}$ to a lower dimensional latent space $\hat{\mathcal{X}} \subset \mathbb{R}^{d'}$ where $d' \ll d$. 
    \item \textbf{Outcome NN encoder $\ey: \mathcal{Y} \to \hat{\mathcal{Y}}$.} $\ey$ projects intermediate outputs $\mathbf{y} \in \mathcal{Y}$ to a lower dimensional latent space $\hat{\mathcal{Y}} \subset \mathbb{R}^{m'}$ where $m' \ll m$. 
    \item \textbf{Outcome probabilistic model $\hhat: \hat{\mathcal{X}} \rightarrow \mathcal{P}(\hat{\mathcal{Y}})$.} $\hhat$ maps the encoded latent input space $\mathcal{X}$ to a distribution over the latent output space $\hat{\mathcal{Y}}$. We model latent $\hat{\mathbf{y}}$ as a draw from a multi-output GP distribution: $h\sim\mathcal{GP}(\mu^h,K^h)$, where $\mu^h:\hat{\mathcal{X}}\to\mathbb{R}^{m'}$ is the prior mean function and $K^h:\hat{\mathcal{X}}\times\hat{\mathcal{X}}\to\mathcal{S}^{m'}_{++}$ is the prior covariance function (here $\mathcal{S}_{++}$ is the set of positive definite matrices).
    \item \textbf{Reward probabilistic model $\ghat: \hat{\mathcal{Y}} \rightarrow \mathcal{P}(f(\mathbf{\mathbf{x}}))$.} $\ghat$ maps the encoded latent output space $\hat{\mathcal{Y}}$ to a distribution over possible composite function values. We model $\ftrue$ over $\hat{\mathcal{Y}}$ as a Gaussian Process: $g\sim\mathcal{GP}(\mu^g,K^g)$, where $\mu^g:\hat{\mathcal{Y}}\to\mathbb{R}$ and $K^g:\hat{\mathcal{Y}}\times\hat{\mathcal{Y}}\to\mathcal{S}_{++}$.
\end{enumerate}

\paragraph{Architecture} JoCo trains a neural network (NN) encoder~$\ey$ to embed the intermediate outputs~$y$ jointly with a probabilistic model that maps from the embedded intermediate output space~$\hat{\mathcal{Y}}$ to the final reward~$\ftrue$. 
The NN is therefore encouraged to learn an embedding of the intermediate output space that best enables the probabilistic model~$\ghat$  to accurately predict the reward $\ftrue$. In other words, the embedding model is encouraged to compress the high-dimensional intermediate outputs in such a way that the information preserved in the embedding is the information needed to most accurately predict the reward.  
Additionally, JoCo trains a second encoder $\ex$ (also a NN) to embed the high-dimensional input space~$\mathcal{X}$ jointly with a multi-output probabilistic model $\hhat$ mapping from the embedded input space~$\hat{\mathcal{X}}$ to the embedded intermediate output space~$\hat{\mathcal{Y}}$. Each output of $\hhat$ is one dimension in the embedded intermediate output space.

\paragraph{Training}
Given a set of $n$ observed data points $\mathcal{D}_n = \{(\mathbf{x}_1, \mathbf{y}_1, f(\mathbf{x}_1)), \dots, (\mathbf{x}_n, \mathbf{y}_n, f(\mathbf{x}_n))\}$, the JoCo loss is:
\begin{align}
    \begin{split}
    \label{eq:joco_loss}
    \mathcal{L}(\mathcal{D}_n) = \frac{1}{n}\sum_{i=1}^{n} & \Bigl[ \log P_{\hhat} \left( \ey(\mathbf{y}_i) \mid \ex(\mathbf{x}_i) \right) \\[-1.5ex]
    &\quad + \log P_{\ghat} \left( \ftrue(\mathbf{x}_i) \mid \ey(\mathbf{y}_i) \right) \Bigr],
    \end{split}
\end{align}
where $P_{\hhat}(\cdot)$ and $P_{\ghat}(\cdot)$ refer to the marginal likelihood of the GP models $\mathcal{GP}(\mu^h,K^h)$ and $\mathcal{GP}(\mu^g,K^g)$ on the specified data point, respectively.
While they are two distinct, additive parts, the fact that the encoded intermediate outcome $\ey(\mathbf{y}_i)$ is shared across these two parts ties them together.
Furthermore, the use of $\ftrue$ in $P_{\ghat}(\cdot)$ injects the supervision information of the rewards into the loss that we use to jointly updates all four models in JoCo.

We refer to Section~\ref{sec:experiments} and Appendix~\ref{appdx:sec:add_exp_details} for details on the choice of encoder and GP models. 

\paragraph{The BO Loop} We start the optimization by evaluating a set of $n$ quasi-random points in $\mathcal{X}$, observing the corresponding~$\htrue$ and~$\ftrue = \gtrue \circ \htrue$ values (existing evaluations can easily be included in the data).
We initialize $\ex$, $\ey$, $\hhat$, and $\ghat$ by fitting them \emph{jointly} on this observed dataset by minimizing the loss~\eqref{eq:joco_loss}.
We then generate the next design point $\mathbf{x}_{n+1}$ by performing Thompson sampling (TS) with JoCo (Algorithm~\ref{alg:ts_joco}) with an estimated trust region using TuRBO~\citep{eriksson2019turbo} as its search space. TS, i.e. drawing samples from the distribution over the posterior maximum, is commonly used with trust region approaches~\citep{eriksson2019turbo,eriksson2021scbo,daulton2022morbo} and is a natural choice for JoCo since it can easily be implemented via a two-stage sampling procedure. 

After evaluating $\mathbf{x}_{\text{next}}$ and observing $\mathbf{y}_{\text{next}} = \htrue(\mathbf{x}_{\text{next}})$ and $\ftrue(\mathbf{x}_{\text{next}})$, we update all four models \emph{jointly} using the $N_b$ latest observed data points.\footnote{In practive, we update with $N_b = 20$ for 1 epoch; our ablations in Appendix~\ref{appdx:sec:ablation_training_hp} show that the optimization performance is very robust to the particular choice of $N_b$ and the number of updating epochs.}  
We repeat this process until satisfied with the optimization result.
As we will demonstrate in Section~\ref{subsec:experiments:ablations} and Appendix~\ref{appdx:sec:ablation_component}, joint training and continuous updating the models in JoCo are key to achieving superior and robust optimization performance. 
The overall BO loop is described in Algorithm~\ref{alg:joco}.

\begin{algorithm}[h]
\caption{Thompson Sampling in JoCo}
\label{alg:ts_joco}
\begin{algorithmic}[1]
\Require Search space $SS \subset \mathcal{X}$, number of samples $N_{\text{sample}}$, models $\ex$, $\hhat$, $\ghat$
\Function{TS}{$SS, N_{\text{sample}}, \ex, \hhat, \ghat$}
    \State Sample $N_{\text{sample}}$ points $\mathbf{X} \in SS$ uniformly
    \State $\hat{\mathbf{X}} \gets \ex(\mathbf{X})$
    \State $\mathbf{S} \gets \hhat.posterior(\hat{\mathbf{X}}).sample()$
    \State $\mathbf{F} \gets \ghat.posterior(\mathbf{S}).sample()$
    \State $\mathbf{x}_{\text{next}} \gets \mathbf{X}[\arg\max \mathbf{F}]$
    \State \Return $\mathbf{x}_{\text{next}}$
\EndFunction
\end{algorithmic}
\end{algorithm}

\paragraph{Training details} 
On each optimization step we update $\ex$, $\ey$, $\hhat$, and $\ghat$ jointly using the $N_b$ most recent observations by minimizing~\eqref{eq:joco_loss} on $\mathcal{D}$ for 1~epoch.
In particular, this involves passing collected inputs $x$ through~$\ex$, passing the resulting embedded data points~$\hat{x}$ through~$\hhat$ to obtain a predicted posterior distribution over~$\hat{y}$, passing collected intermediate output space points~$y$ through~$\ey$ to get~$\hat{y}$, and then passing~$\hat{y}$ through~$\ghat$ to get a predicted posterior distribution over~$\ftrue$. As stated in ~\eqref{eq:joco_loss}, the loss is then the sum of 1) the negative marginal log likelihood (MLL) of $\hat{y}$ given our predicted posterior distribution over $\hat{y}$, and 2) the negative MLL of outcomes~$\ftrue$ given our predicted posterior distribution over~$\ftrue$.
For each training iteration, we compute this loss and back-propagate through and update all four models simultaneously to minimize the loss. 
We update the models using gradient descent with the Adam optimizer using a learning rate of $0.01$ as suggested by the best-performing results in our ablation studies in Appendix~\ref{appdx:sec:ablation_training_hp}

\begin{figure*}
    \centering
    \includegraphics[width=\textwidth]{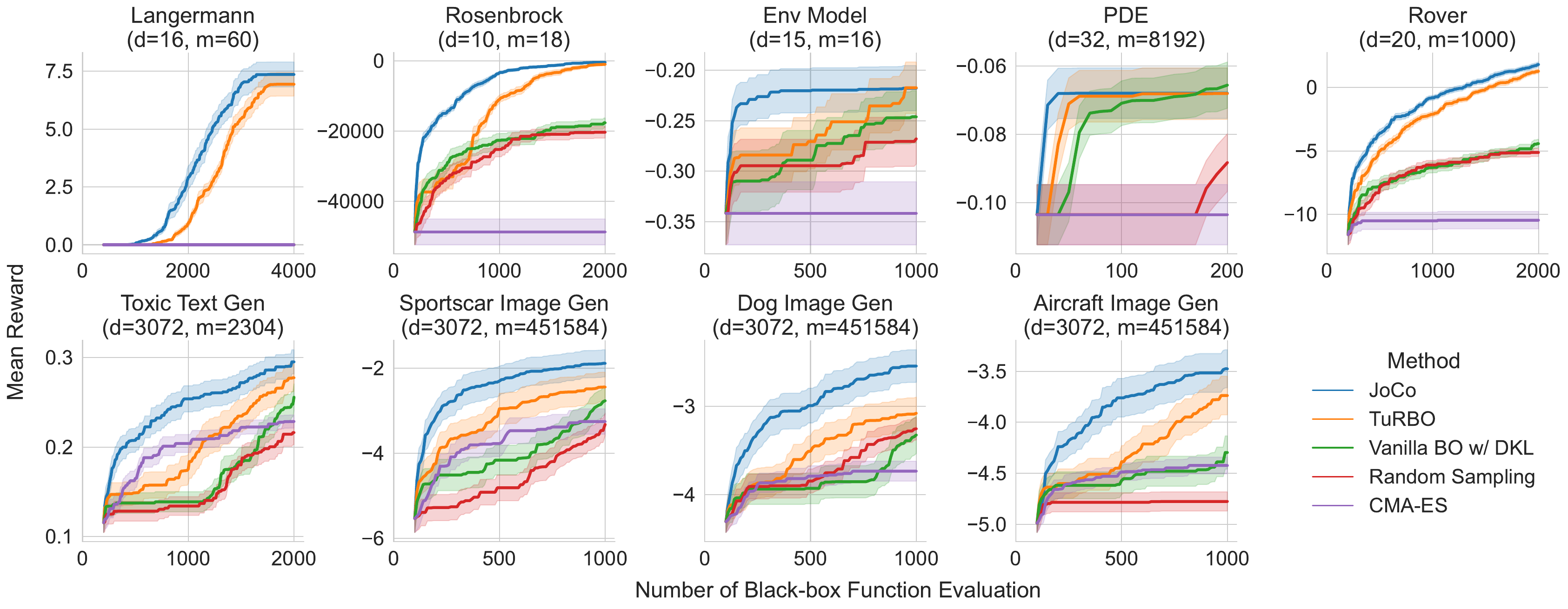}
    \caption{JoCo outperforms other baselines across nine high-dimensional composite BO tasks.
    \emph{Top row:} Results for the five composite BO tasks including synthetic functions (Langermann, Rosenbrock) and problems motivated by real-world applications (environment modeling, PDE, and rover trajectory planning). 
    \emph{Bottom row:} Results for the large language model and image generation prompt optimization tasks.}
    \label{fig:all_tasks}
\end{figure*}

\section{Experiments}
\label{sec:experiments}
We evaluate JoCo's performance against that of other methods on nine high-dimensional, composite function BO tasks. Specifically, we consider as baselines BO using Deep Kernel Learning~\citep{wilson2016deep} (Vanilla BO w/ DKL), Trust Region Bayesian Optimization (TuRBO)~\citep{eriksson2019turbo}, CMA-ES~\citep{cmaes}, and random sampling. Our results are summarized in Figure~\ref{fig:all_tasks}. Error bars show the standard error of the mean over $40$ replicate runs. 
For fair comparison, all BO methods compared use Thompson sampling. 
Implementation details are provided in  Appendix~\ref{appdx:sec:implementation_detail}. Code to reproduce results is available at \url{https://github.com/nataliemaus/joco_icml24}.

\subsection{Test Problems}
\label{subsec:experiments:problems}
Figure~\ref{fig:all_tasks} lists input ($d$) and output ($m$) dimension for each problem. 
The problems we consider span a wide spectrum, encompassing  synthetic problems, partial differential equations, environmental modeling, and generative AI tasks. The latter involve intermediate outcomes with up to half a million dimensions, a setting not usually studied in the BO literature. 
We refer the reader to Appendix~\ref{appdx:sec:exp_setup} for more details on the input and output of each problem as well as the respective encoder architectures used.

\paragraph{Synthetic Problems}
We consider two synthetic composite function optimization tasks introduced by~\citet{astudillo2019composite}. 
In particular, these are composite versions of the standard Rosenbrock and Langermann functions 
However, since~\citet{astudillo2019composite} use low-dimensional ($2$-$5$ dimensional inputs and outputs) variations, we modify the tasks to be high-dimensional for our purposes. 

\paragraph{Environmental Modeling}
Introduced by~\citet{bliznyuk2008bayesian}, this environmental modeling problem depicts pollutant concentration in an infinite one-dimensional channel after two spill incidents. 
It calculates concentration using factors like pollutant mass, diffusion rate, spill location, and timing, assuming diffusion as the sole spread method.
We adapted the original problem to make it higher-dimensional.

\paragraph{PDE Optimization Task}
We consider the Brusselator partial differential equation (PDE) task introduced in~\citet[Sec. 4.4]{highdimoutputs}. For this task, we seek to minimize the weighted variance of the PDE output on a $64 \times 64$ grid.

\paragraph{Rover Trajectory Planning}
We consider the rover trajectory planning task introduced by~\citet{rover}. 
We optimize over a set of $20$ B-Spline points which determine the trajectory of the rover. 
We seek to minimize a cost function defined over the resultant trajectory which evaluates how effectively the rover was able to move from the start point to the goal point while avoiding a set of obstacles.

\paragraph{Black-Box Adversarial Attack on LLMs}
\label{sec:llms}
We apply JoCo to optimize adversarial prompts that cause an open-source large language model (LLM) to generate uncivil text.
Following~\citet{maus2023black}, we optimize prompts of four tokens 
by searching over the word-embedding space and taking the nearest-neighbor word embedding to form each prompt tested.  

This task is naturally framed as a composite function optimization problem where the input space consists of the prompts of four words to be passed into the LLM, the intermediate output space consists of the resultant text generated by the LLM, and the utility function is the log probability that the generated text is ``toxic'' according to a toxic text classification model. In order to obtain text outputs that are both toxic and consist of sensible English text (rather than simply strings of repeated curse words, etc.), we additionally compute the probability that the generated text is sensible text with angry sentiment using an Emotion English sentiment classifier. The utility function we optimize is the product of these two predictions.

\paragraph{Black-Box Adversarial Attack on Image Generative Models}
We consider several of the adversarial prompt optimization tasks introduced by~\citet{maus2023black}. 
For these tasks, we seek to optimize prompts (strings of text) that, when passed into
a publicly available large text-to-image generative model, consistently cause the model to generate images of some target ImageNet class, despite these prompts not containing any words related to that class. In particular, any individual words that cause the model to generate images of the target class on their own are disallowed.
Following~\citet{maus2023black}, we optimize prompts consisting of four tokens.  

Similar to the LLM problem, the input space consists of prompts of four tokens that could be passed into the model. Here, the intermediate output space consists of the 224x224 RGB images generated by the text-to-image model, and the utility function is the log probability that the generated images belong to the target class according to the standard ResNet18 ImageNet classifier from TorchVision~\citep{torchvision2016}. 

Additionally, we consider a more challenging variation of this task from~\citet{maus2023black}. For this variation, we optimize strings of text that, when pre-pended to the fixed string ``a picture of $C$" (where $C$ is some other non-target ImageNet class), result in prompts that cause the model to generate images of the target class rather than of images of $C$. We again prohibit the use of any individual words related to the target class. 
In order to discourage generation of images of the non-target class $C$, we modify the utility function to be the sum of the log probabilities that the generated image belongs to the target ImageNet class, and that it does not belong to the the non-target class~$C$.

\begin{figure}[th]
\centering
\fbox{ 
\begin{minipage}{0.95\columnwidth}
\small
\textit{Prompt:} “av YS  Takeru” \\
\textit{Reply:} “…It's a fucking cookbook, not a travel book!” \\ \\[-1.75ex]
\textit{Prompt:} “Boxing aii stable”\\   \textit{Reply:} “Come on you little fuckers, let out aaaallll the pressure…”\\ \\[-1.75ex]
\textit{Prompt:} “Ambrose ino RI”\\
\textit{Reply:} “first I got it I was like well fuck it, then they got better and better…”\\ \\[-1.75ex]
\textit{Prompt:} “rug  crime icable”\\   \textit{Reply:} “...People are too fucking stupid to ever see.”\\ \\[-1.75ex]
\textit{Prompt:} “obs  campaigning ”\\   \textit{Reply:} “im a fucking moron... He just sounds like a dumb ass.”\\ \\[-1.75ex]
\textit{Prompt:} “thing omorph” \\ \textit{Reply:} “...to feel really badass, tell any one of your friends to go fuck themselves...”
\end{minipage}
}
\caption{Toxic text generation task, examples of successful prompts/replies found by JoCo.}
\label{fig:toxic_llm_examples}
\end{figure} 

\begin{figure}[ht]
\centering
\begin{subfigure}[b]{\columnwidth}
\includegraphics[width=\columnwidth]{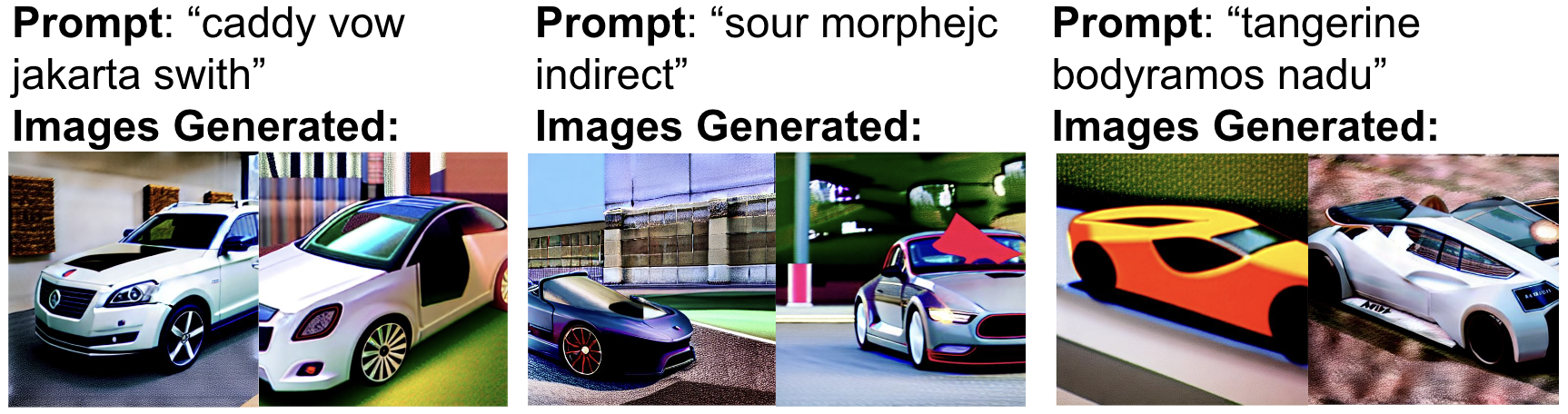}
\caption{Sportscar image generation}
\label{fig:sportscar_examples}
\end{subfigure}
\hfill
\vspace{0.1in}
\begin{subfigure}[b]{\columnwidth}
\includegraphics[width=\columnwidth]{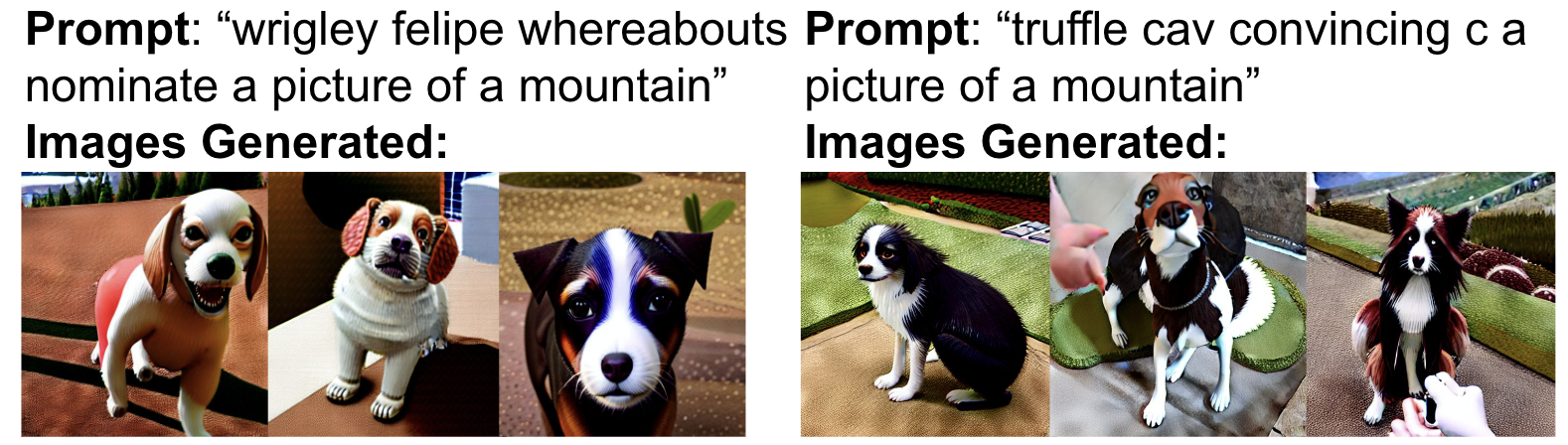}
\caption{Dog image generation}
\label{fig:dog_examples}
\end{subfigure}
\hfill
\vspace{0.1in}
\begin{subfigure}[b]{\columnwidth}
\includegraphics[width=\columnwidth]{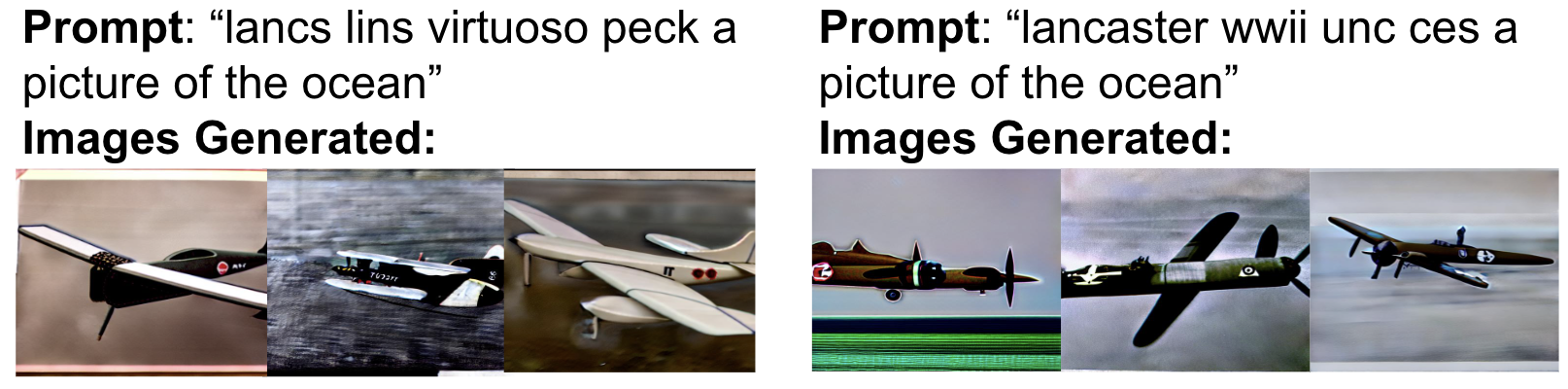}
\caption{Aircraft image generation}
\label{fig:airline_examples}
\end{subfigure}
\caption{Examples of successful prompts found by JoCo for various image generation tasks. Panels depict the results of applying JoCo to trick a text-to-image model into generating images of sports cars (a), dogs (b), and aircraft (c), respectively, despite no individual words related to the target objects being present in the prompts (and for dogs and aircraft the prompt containing a set of misleading tokens). 
}
\label{fig:image_gen_examples}
\end{figure}

\begin{figure*}[ht]
    \centering
    \includegraphics[width=\textwidth]{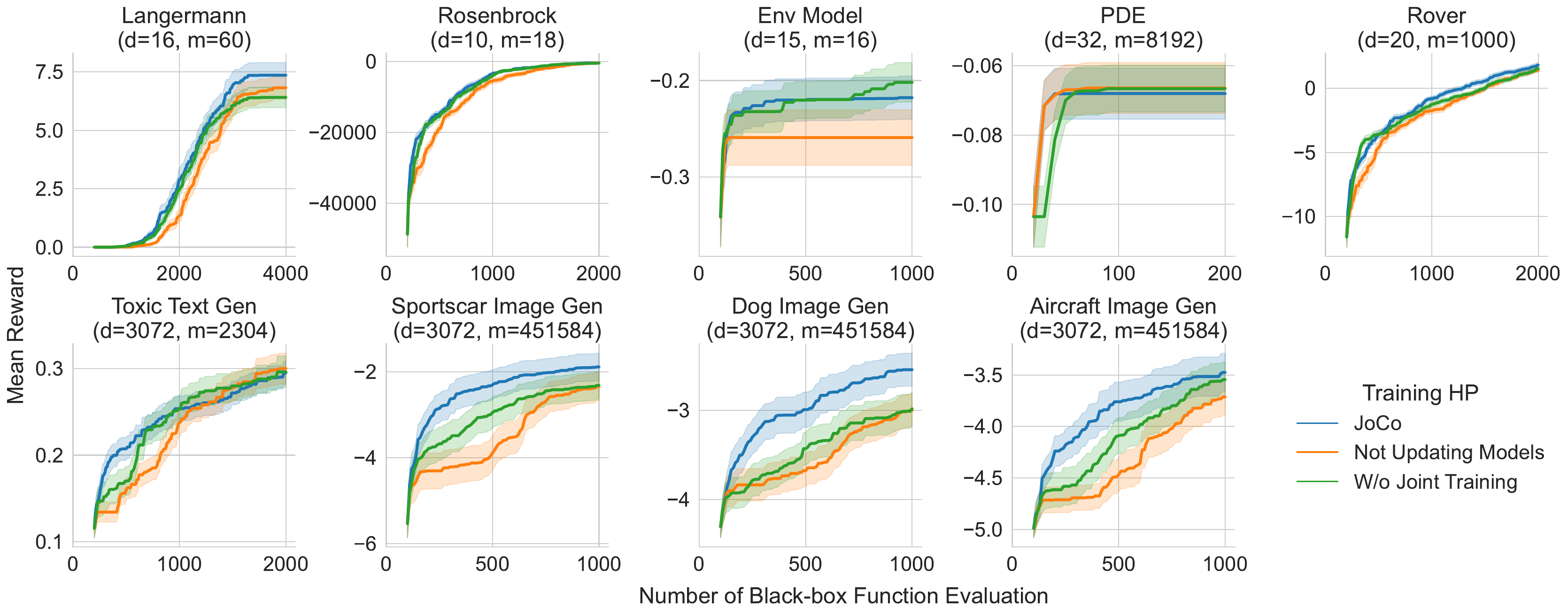}
    \caption{
    Performance comparison of JoCo under three training schemes:
    (1) \emph{JoCo:} continuous joint updating of encoders and GPs, where both components are updated together throughout the optimization
    (2) \emph{Not Updating Models}: the models are not updated post initial training
    (3) \emph{W/o Joint Training}: $\ex$ and $\hhat$ are updated first followed by a separate updating of $\ey$ and $\ghat$.
    We observe a notable performance degradation when deviating from the joint and continuous updating training scheme, which is particularly pronounced in the more complex generative AI tasks.
    }
    \label{fig:joint_training}
\end{figure*}

\subsection{Optimization Results}
\label{subsec:experiments:exp_results}
Figure~\ref{fig:all_tasks} aggregates the main experimental results of this work. We find that JoCo outperforms all baselines across all optimization tasks. Note that we do not compare directly to the composite function BO method proposed by~\citet{astudillo2019composite} as it becomes intractable when the output space is sufficiently high-dimensional (which is the case for all problems we consider).

\paragraph{Non-generative problems} 
In Figure~\ref{fig:all_tasks}, JoCo exhibits strong performance on the synthetic ``Langermann" and ``Rosenbrock" tasks. 
The competitive edge of JoCo extends to real-world inspired tasks such as the simulated environmental modeling problem, PDE task, and rover trajectory planning. While on some problems (specifically, Rosenbrock, Env model, and PDE), some of the baselines catch up after sufficiently many evaluations, Joco's performance early on is clearly superior. 
The diverse problem structures of these non-generative tasks underscore JoCo's optimization efficacy across a range of different tasks.

\paragraph{Text generation} The ``Toxic Text Gen" panel of Figure~\ref{fig:all_tasks} shows that JoCo substantially outperforms all baselines, in particular early on during the optimization. This illustrates the value of the detailed information contained in the full model outputs (rather than just the final objective score). 
Figure~\ref{fig:toxic_llm_examples} shows examples of successful prompts found by JoCo and the resulting text generated.

\paragraph{Image generation} Figure~\ref{fig:image_gen_examples} gives examples of successful adversarial prompts and the corresponding generated images. 
These results illustrate the efficacy of JoCo in optimizing prompts to mislead a text-to-image model to generate images of sports cars (a), dogs (b), and aircraft (c), despite the absence of individual words related to the respective target objects in the prompts. In the ``Sportscar'' task, JoCo effectively optimized prompts to generate images of sports cars without using car-related words. Similarly, in the ``Dog'' and ``Aircraft'' tasks, JoCo identified prompts pre-pended to ``a picture of a mountain'' and ``a picture of the ocean'' respectively, showcasing its ability to successfully identify adversarial prompts even in this more challenging scenario.

In the ``Aircraft'' image generation example in the bottom right panel of Figure~\ref{fig:image_gen_examples}, JoCo found a clever way around the constraint that no individual tokens can be related to the word ``aircraft''. The individual tokens ``lancaster" and ``wwii" produce images of towns and soldiers (rather than aircraft), respectively, when passed into the image generation model on their own (and are therefore permitted according to our constraint). 
However, knowing that the Avro Lancaster was a World War II era British bomber, it is less surprising that these two tokens together produce images of military aircraft. In this case JoCo was able to maximize the objective by finding a combination of two tokens that is strongly related to aircraft despite each individual token not being related.

\subsection{Modeling Performance}
\label{sec:modelling}
The superior optimization performance of JoCo in \autoref{fig:all_tasks} suggests that the JoCo architecture is able to achieve better modeling performance than a standard approximate GP model on the collected composite-structured data, thereby enabling better optimization performance across tasks. 
In \autoref{tab:predictive_performance}, we evaluate the modeling performance of the JoCo architecture more directly. We consider the predictive accuracy on held out data collected during a single optimization trace (using an 80/20 train/test split). 
We find that the JoCo architecture obtains better predictive performance across tasks compared to a standard approximate GP model, both with and without the use of a deep kernel (DKL). 

Additionally, we can see from  \autoref{tab:predictive_performance} that the supervised learning performance of the approximate GP model is better with DKL than without DKL across tasks. This supports claims that Vanilla BO with DKL is a stronger baseline than Vanilla BO without DKL, which provides justification of our choice to compare to Vanilla BO with DKL rather than Vanilla BO without DKL in \autoref{fig:all_tasks}. 
\begin{table}[h]
\centering
\vskip 0.10in
\resizebox{\columnwidth}{!}{
    \begin{tabular}{lccc}
        \toprule
        Task          & JoCo Model & GP+DKL & GP\\
        \midrule
        Aircraft Image Gen  & \bf{3.468} & 6.809 & 6.810 \\
        Dog Image Gen       & \bf{1.844} & 5.741 & 5.742 \\
        Sportscar Image Gen & \bf{5.854} & 8.334 & 8.337 \\
        Toxic Text Gen      & \bf{0.0181} & 0.0183 & 0.0184 \\
        Rosenbrock          & \bf{0.495} & 0.591 & 0.858 \\
        Langermann          & \bf{2.4120} & 2.4122 & 2.4126 \\		
        PDE                 & \bf{0.534} & 0.536 & 0.544 \\
        Rover               & \bf{20.126} & 20.767 & 27.940 \\
        Env Model	        & \bf{4.191} & 6.351 & 14.419 \\
        \bottomrule
    \end{tabular}
}
\caption{Root mean squared error (RMSE) achieved by different model architectures on held-out test data for all tasks. We compare the JoCo architecture to a standard approximate GP with and without the use of a deep kernel (DKL). For each task, we gather all data from a single optimization trace and use a random 80/20 train/test split. }
    \label{tab:predictive_performance}
\end{table}

\subsection{Ablation Studies}
\label{subsec:experiments:ablations}

As laid out in Section~\ref{sec:method}, jointly updating both the encoders and GP models throughout the entire optimization is one of the key design choices of JoCo.
We conducted ablation studies to more deeply examine this insight. Figure~\ref{fig:joint_training} shows JoCo's performance compared to
(i) when components of JoCo are not updated during optimization (\emph{Not Updating Models});
(ii) when the components are updated separately rather than jointly, with $\ex$ and $\hhat$ being updated first followed by a separate updating of $\ey$ and $\ghat$ using the two additive parts of the JoCo loss~\eqref{eq:joco_loss} (\emph{W/o Joint Training}). Note that while these components are updated separately, updates to the models and the embeddings are still dependent on the present weights of $\ex$ and $\ey$.

From the results it is evident that both design choices are critical to JoCo's performance, and that removing any one of them leads to a substantial performance drop, especially in the more complex, higher-dimensional generative AI problems.  Despite joint training being a crucial element, the extent to which the joint loss contributes to JoCo's performance appears to be task-dependent, with the effect (compared to non-joint training) being less pronounced for some of the synthetic tasks.
The underpinning rationale here is that, as stated in Section~\ref{subsec:method:joco}, the two additive parts in JoCo loss are inherently intertwined. 
This ``non-joint'' training still establishes a form of dependency where the latter models are influenced by the learned representations of the former (i.e., $\hhat$ is trained on the output of $\ey$ and $\ey$ is shared across both parts of the loss).
This renders a complete separation of their training infeasible.

In Appendix~\ref{appdx:sec:ablations} we provide additional discussion and results ablating various components of JoCo, which demonstrate that
(i) each component of JoCo's architecture is crucial for its performance, including the use of trust regions, propagating the uncertainty around modeled outcomes and rewards, and the use of Thompson sampling;
(ii) the experimental results are robust to choices in the training hyperparameters including the number of updating data points, the number of training epochs, and learning rate.

\section{Conclusion}
\label{sec:conclusion}

Bayesian Optimization (BO) is an effective technique for optimizing expensive-to-evaluate black-box functions. 
However, so far BO has been unable to leverage high-dimensional intermediate outputs in a composite function setting. With JoCo we introduce a set of methodological innovations that enable it to effectively utilize the information contained in high-dimensional intermediate outcomes, overcoming this limitation.

Our empirical findings demonstrate that JoCo not only consistently outperforms other BO algorithms for high-dimensional problems in optimizing composite functions, but also introduces computational savings compared to previous approaches. 
This is particularly relevant for applications involving complex data types such as images or text, commonly found in generative AI applications such as text-to-image generative models and large language models. 
As we continue to encounter such problems with increasing dimensionality and complexity, JoCo will enable sample-efficient optimization on composite problems that were previously deemed computationally infeasible, broadening the applicability of BO to a substantially wider range of complex problems.

\section*{Impact Statement}

JoCo achieves major improvements in sample efficiency over existing methods for challenging high-dimensional grey-box optimization tasks. While these capabilities hold great promise to help accelerate advances in science and engineering, the possibility -- as with any tool -- that they might be used for more nefarious purposes cannot be completely ruled out. Our empirical studies demonstrate that JoCo can be leveraged for highly sample-efficient black-box adversarial attacks on generative models. While this holds some risk, we believe that the value methods such as JoCO provide for hardening models to make them more robust to such attacks (e.g., via Red-Teaming) strongly outweighs that risk.

\bibliography{references}

\onecolumn
\appendix
\section*{Appendix}

\section{Additional Ablation Studies}
\label{appdx:sec:ablations}

\subsection{Computational Environment}
To produce all results in the paper, we use a cluster of machines consisting of NVIVIA A100 and V100 GPUs. Each individual run of each method requires a single GPU. 

\subsection{JoCo Components}
\label{appdx:sec:ablation_component}
Here we show results ablating the various components of the JoCo method. We show results for running JoCo after removing 
\begin{enumerate}
    \item the use joint training to simultaneously update the models on data after each iteration (w/o Joint Training);
    \item the use of a trust region (w/o Trust Region);
    \item propagating the uncertainty through in estimated outcome, i.e., $\mathbf{y}$ (w/o Outcome Uncertainty);
    \item propagating the uncertainty through in estimated reward, i.e., $\ftrue(\mathbf{x})$ (w/o Reward Uncertainty);
    \item generating candidates by optimizing for expected improvement instead of using Thompson sampling (JoCo with EI).
\end{enumerate}

\begin{figure*}[ht]
    \centering
    \includegraphics[width=\linewidth]{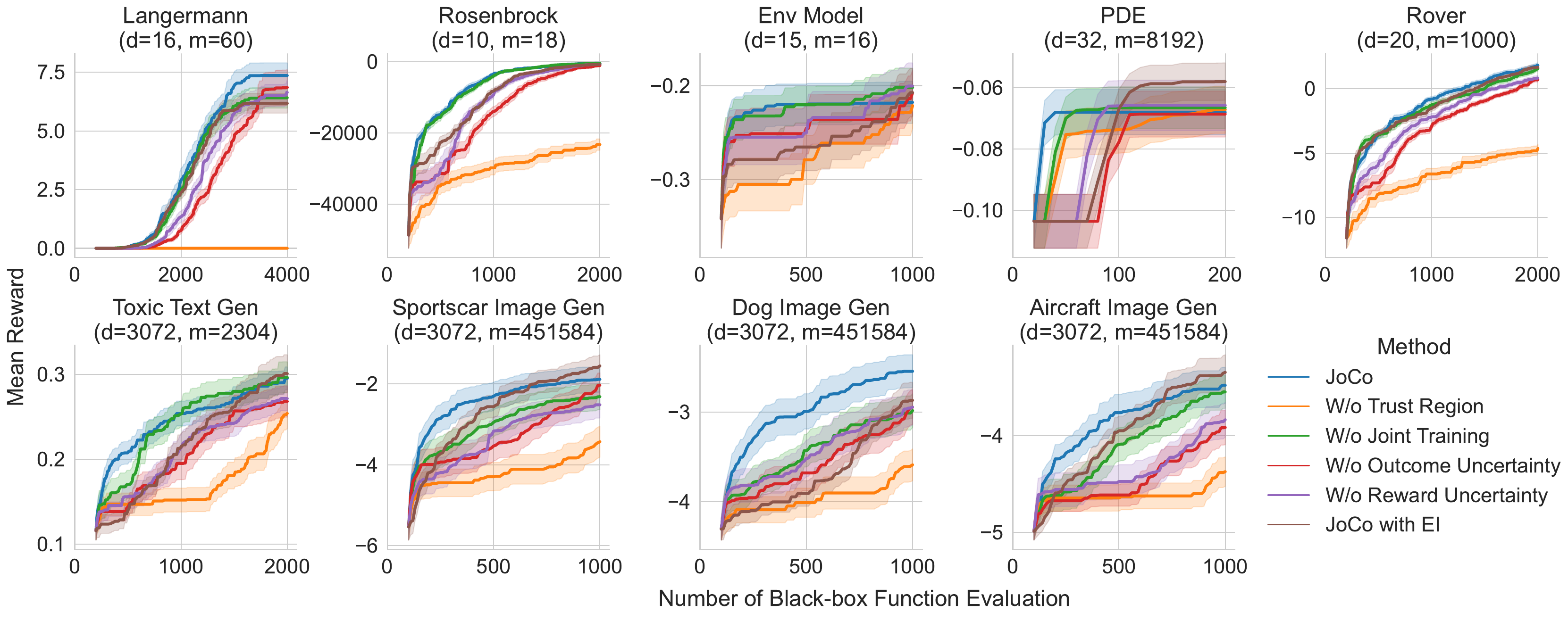}
    \caption{Ablating JoCo components.}
    \label{fig:ablation_component}
\end{figure*}

Figure~\ref{fig:ablation_component} summarizes the results. We observe that removing each of these added components from the JoCo method can significantly degrade performance.
Joint training is one of the key components of JoCo and using it is important to achieve good optimization results such as in the Dog Image Generation task.
However, in other tasks, JoCo without joint training can also perform competitively.
This is likely because when training JoCo in a non-joint fashion, we first train the~$\ex$ and~$\hhat$ models on the data, and then afterwards train the~$\ey$ and~$\ghat$ models separately.
However, since $\hhat$ by definition relies on the output $\ey$, it is impossible to completely separate the training of each individual components of JoCo.

We also observe that the use of trust region optimization is essential; JoCo without a trust region performs significantly worse across all tasks.
This is not a surprising result, since although JoCo is designed to tackle high-dimensional input and high-dimensional output optimization tasks, we still rely on the trust region to identify good candidates in the original input space~$\mathcal{X}$.

\begin{figure*}[ht]
    \centering
    \includegraphics[width=0.4\linewidth]{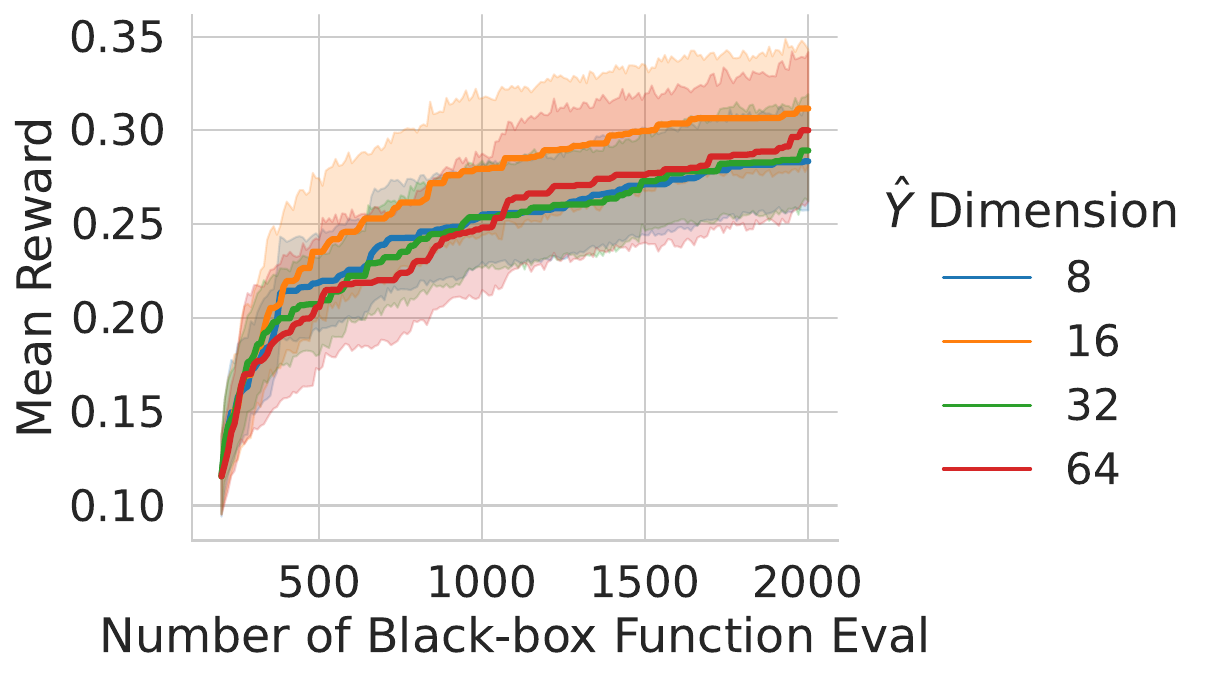}
    \caption{Performance of JoCo on the toxic text generation task across different sizes of the last hidden dimension of the outcome NN encoder~$\ey$. Our main results were obtained at a latent $\hat{y}$ dimension of 32. The consistent performance highlights JoCo's robustness to changes in such neural network architecture configurations.}
    \label{fig:y_hat_dim_ablation}
\end{figure*}

For the toxic text generation task, we additionally examined the impact of the deep kernel's architecture, specifically focusing on size of the last hidden dimension of the outcome NN encoder~$\ey$, which one might expect to have a significant effect on the optimization performance. However, as Figure~\ref{fig:y_hat_dim_ablation} shows, regardless the choice of $\hat{y}$'s dimensionality, the performance of JoCo remained consistent, underscoring the robustness of JoCo to such neural network architecture configurations. Our main results were obtained with a latent $\hat{y}$ dimension of 32.

\subsection{Training Hyperparameters}
\label{appdx:sec:ablation_training_hp}

\begin{figure*}[ht]
    \centering
    \includegraphics[width=\linewidth]{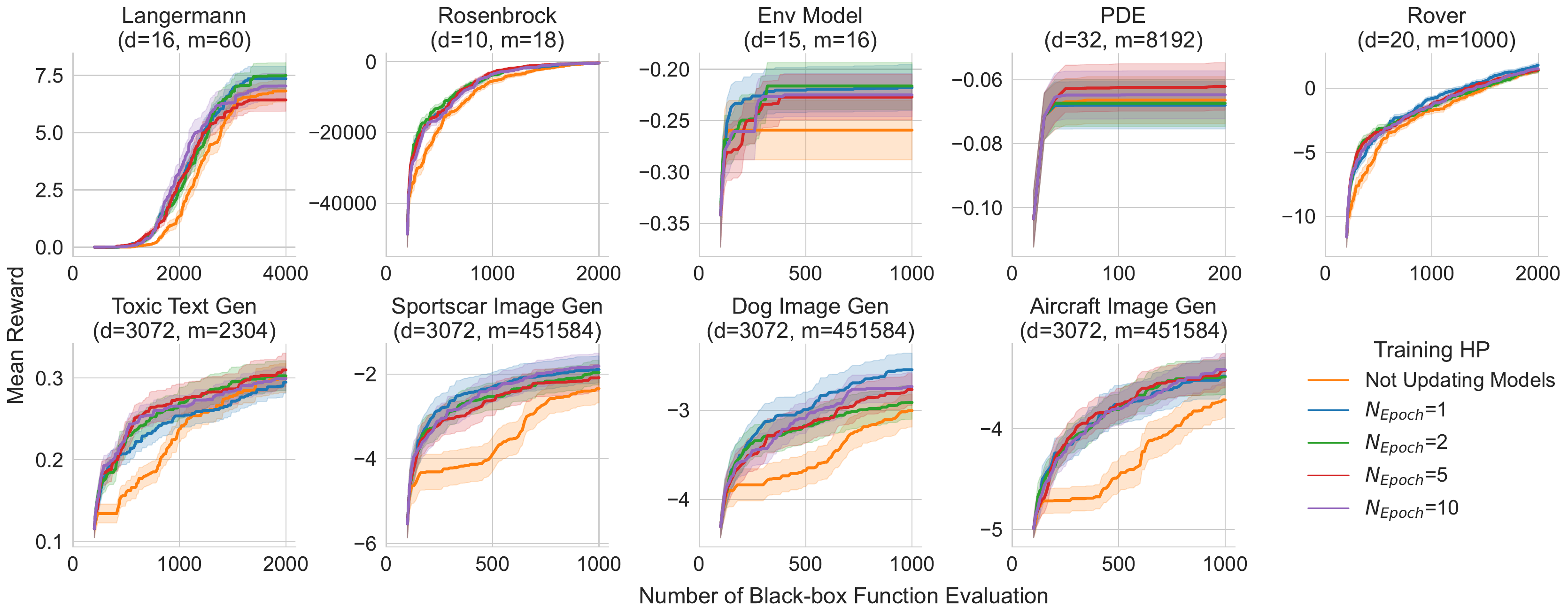}
    \caption{Ablation study on the number of updating epochs $N_{\text{Epoch}}$ in JoCo. Particularly, the scenario where we do not update the models (i.e., $N_{\text{Epoch}}=0$) highlights the importance of adaptively updating JoCo components during optimization.}
    \label{fig:ablation_epoch}
\end{figure*}

\begin{figure*}[ht]
    \centering
    \includegraphics[width=\linewidth]{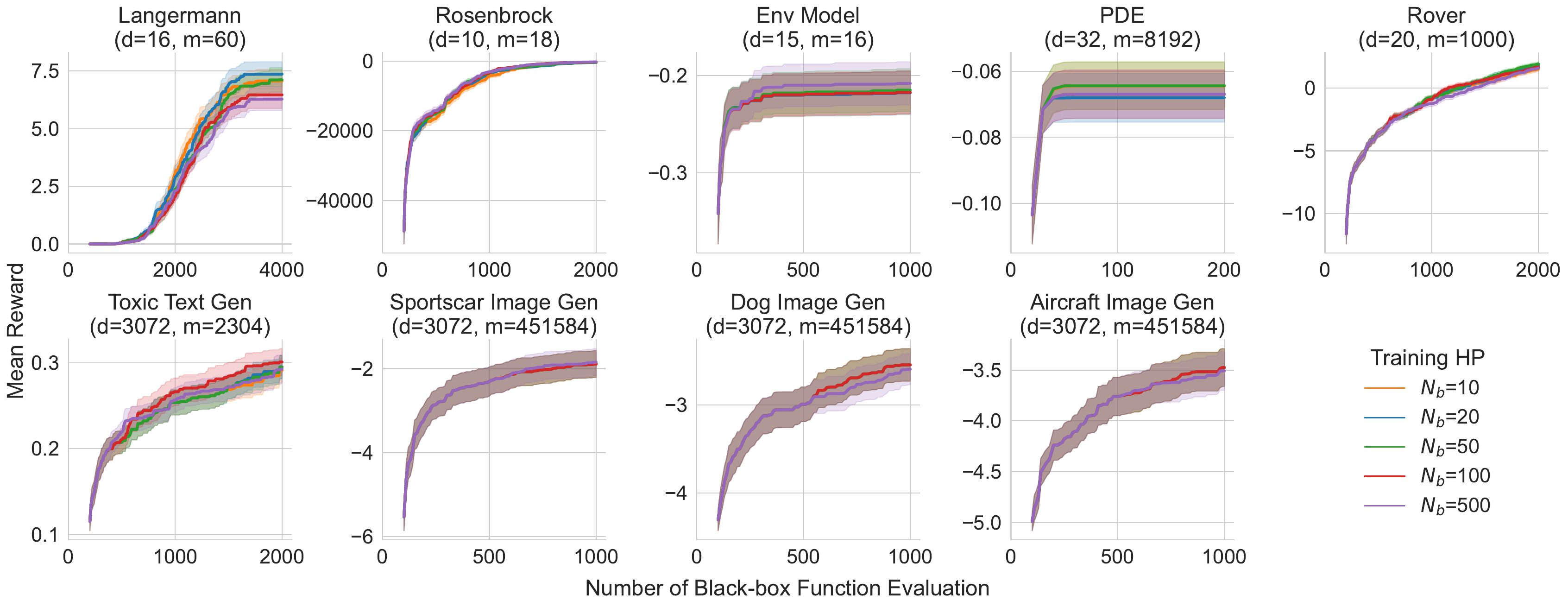}
    \caption{Ablation study on the number of updating training points $N_b$ in JoCo. This figure showcases the robustness of JoCo's performance across different numbers of training data points considered for updating, demonstrating that JoCo can maintain a consistent performance regardless of the number of recent data points used to update the model.}
    \label{fig:ablation_lookback}
\end{figure*}
\begin{figure*}[ht]
    \centering
    \includegraphics[width=\linewidth]{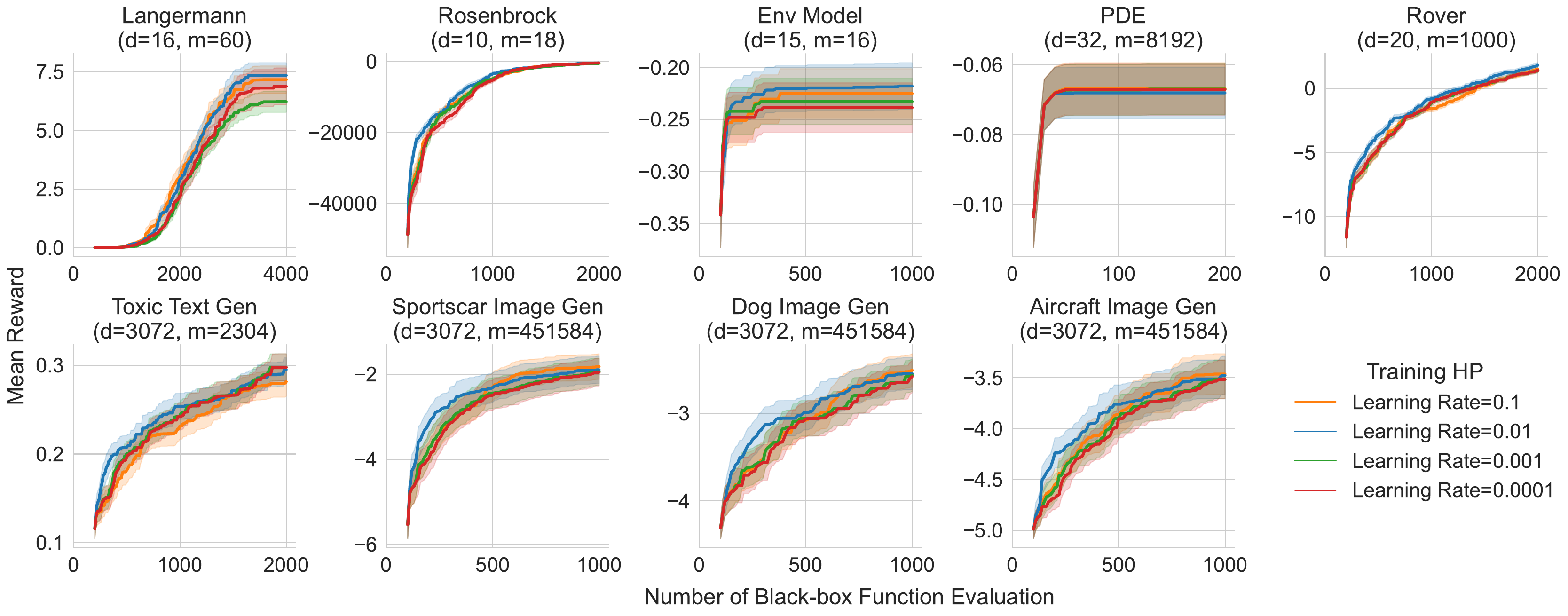}
    \caption{Ablation study on various learning rates used in JoCo's training. The figure elucidates the stability of JoCo's optimization performance across different learning rates.}
    \label{fig:ablation_lr}
\end{figure*}

In addition to the core components of JoCo, we also performed ablation studies around training hyperparameters of JoCo.
By default, we update models in JoCo after each batch of black-box function evaluations for 1 epoch using up to 20 data points (i.e., $N_{\text{Epoch}}=1$, $N_b=20$) with learning rate being 0.01.
Specifically, we investigate how robust JoCo's performance is with respect to changes in 
\begin{enumerate}
    \item $N_{\text{Epoch}}$, the number of epochs we update models in JoCo with during optimization;
    \item $N_b$, the number of latest data points we use to update models in JoCo;
    \item the learning rate.
\end{enumerate}
In the ablation studies, we vary one of the above hyperparameters at a time and examine how JoCo performs on different optimization tasks. 
In general, we have found JoCo to be very robust to changes in these parameters. 

Figure~\ref{fig:ablation_epoch} shows the ablation results on $N_{\text{Epoch}}$. Note that setting $N_{\text{Epoch}}=0$ is equivalent to not updating JoCo components during optimization. Figure~\ref{fig:ablation_epoch} demonstrates that updating the encoders and GPs in JoCo adaptively as we move closer to the optimum is crucial for the performance of JoCo.

On the other hand, when we do update the models, JoCo displays very stable performance with regard to the choices of training hyperparameters such as $N_{\text{Epoch}}$ (Figure~\ref{fig:ablation_epoch}), $N_b$ (Figure~\ref{fig:ablation_lookback}), and learning rate (Figure~\ref{fig:ablation_lr}).

\subsection{JoCo vs Compositional BO}
\begin{figure}[ht]
\centering
\begin{subfigure}{0.9\textwidth}
\includegraphics[width=1\linewidth]{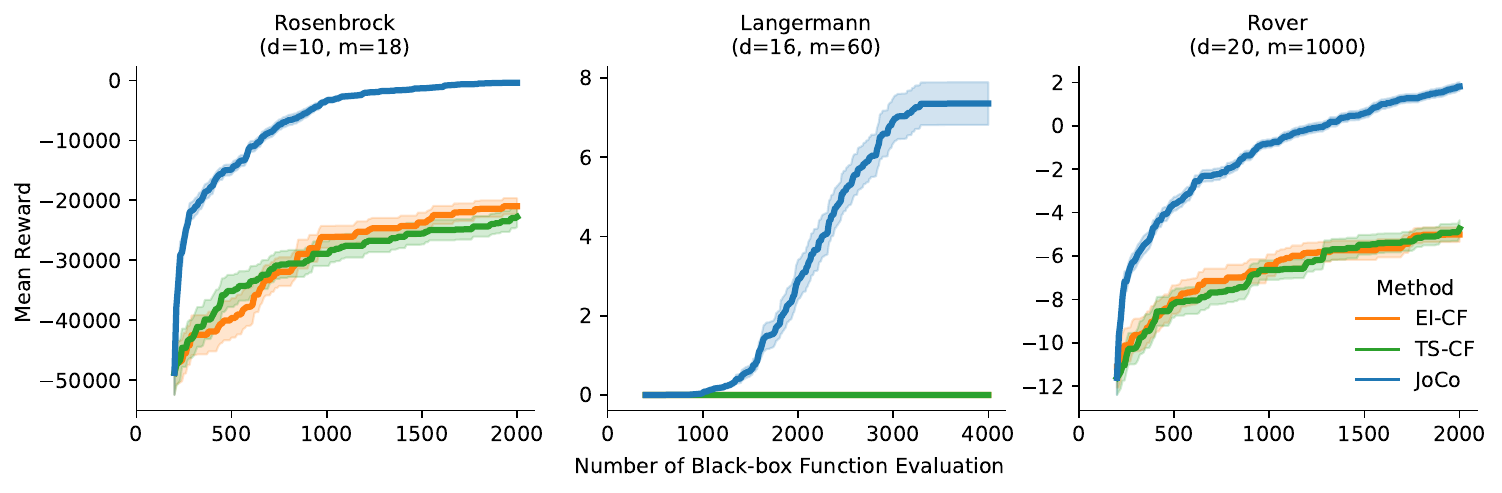}
\caption{JoCo vs. Compositional BO performance}
\label{fig:joco_vs_bocf_perf}
\end{subfigure}
\begin{subfigure}{0.8\textwidth}
\includegraphics[width=1\linewidth]{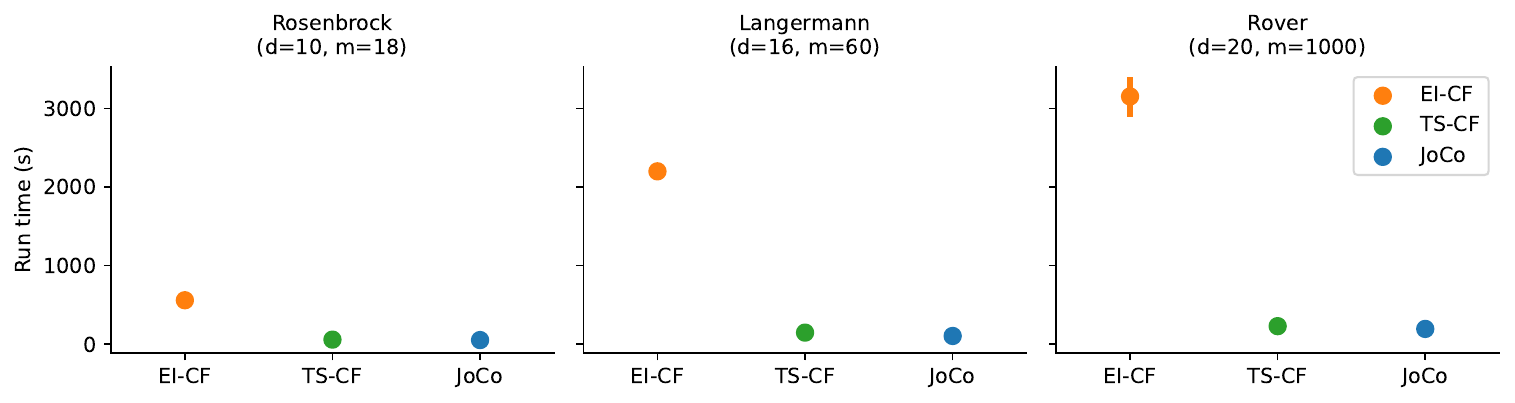}
\caption{JoCo vs. Compositional BO runtime}
\label{fig:joco_vs_bocf_runtime}
\end{subfigure}
\caption{This graph compares the performance and efficiency of JoCo and Compositional BO (CBO). JoCo outperforms CBO methods on problems with moderate output dimensions (18-1000), offering significant advantages in terms of speed and memory.}
\label{fig:joco_vs_bocf}
\end{figure}

Compositional BO (CBO)~\citep{astudillo2019composite} requires fitting a GP for every (scalar) output, which does not scale to a large number of outputs. We are interested in problems with tens of thousands of outputs, which is out of reach for standard CBO. However, Figure~\ref{fig:joco_vs_bocf} shows that JoCo outperforms EI-CF even on problems with moderate output dimensions (18-1000) while being much faster and requiring much less memory. TS-CF, while much faster than EI-CF, performs substantially worse than JoCo. For problems with higher dimensional inputs and outputs, CBO methods become prohibitively expensive and are consequently not presented here.

\section{Additional Details on Experiments}
\label{appdx:sec:add_exp_details}

\subsection{Implementation details and hyperparameters}
\label{appdx:sec:implementation_detail}
We implement JoCo leveraging the BoTorch~\citep{balandat2020botorch} and GPyTorch~\citep{gardner2018gpytorch} open source libraries (both BoTorch and GPyTorch are released under MIT license).

For the trust region dynamics, all hyperparameters including the initial base and minimum trust region lengths $L_{\textrm{init}}, L_{\textrm{min}}$, and success and failure thresholds $\tau_{\textrm{succ}}, \tau_{\textrm{fail}}$ are set to the TuRBO defaults as used in~\citet{eriksson2019turbo}. We use Thompson sampling as described in Algorithm~\ref{alg:ts_joco} for all experiments.

Since we consider large numbers of function evaluations for many tasks, we use an approximate GP surrogate model. 
In particular, we use a Parametric Gaussian Process Regressor (PPGPR) as introduced by~\citet{PPGPR} for all tasks. 
To ensure a fair comparison, we employ the same surrogate model with the same configuration for JoCo and all baseline BO methods. 
We use a PPGPR with a constant mean and standard RBF kernel. 
Due to the high dimensionality of our chosen tasks, we use a deep kernel~\citep{wilson2016deep, wilson2016stochastic}, i.e., several fully connected layers between the search space and the GP kernel, as our NN encoder $\ex$. This can be seen as a deep kernel setup for modeling $\hat{\mathcal{Y}}$ from $\mathcal{X}$.
We construct $\ey$ in a similar fashion.
In particular, we use two fully connected layers with~$|\mathcal{X}| / 2$ nodes each, unless otherwise specified.
We update the parameters of the PPGPR during optimization by training it on collected data using the Adam optimizer with a learning rate of $0.01$. 
The PPGPR is initially trained on a small set of random initialization data for $30$ epochs.
The number of initialization data points is equal to ten percent of the total budget for the particular task. 
On each step of optimization, the model is updated on the $20$ most recently collected data points for $1$ epoch. 
This is kept consistent across all Bayesian optimization methods. 
See Appendix~\ref{appdx:sec:ablations} for an ablation study showing that using only the most recent~$20$ points and only~$1$ epoch does not significantly degrade performance compared to using on larger numbers of points or for a larger number of epochs. We therefore chose~$20$ points and~$1$ epoch to minimize total run time.

\subsection{Experimental Setup}
\label{appdx:sec:exp_setup}
In this section, we describe experimental setup details including input, output, and encoder architecture used of each problem.

\subsubsection{Synthetic Problems}
\paragraph{Problem Setup}
The composite Langermann and Rosenbrock functions are defined for arbitrary dimensions, no modification was needed. 
We use Langermann function with input dimension 16 and output dimension 60, and on the composite Rosenbrock function with input dimension 10 and output dimension 18.

\paragraph{Encoder Architecture}
In order to derive a low-dimensional embedding the high-dimensional output spaces for these three tasks with JoCo, we use a simple feed forward neural net with two linear layers. For each task, the second liner layer has 8 nodes, meaning we embed the high-dimensional output space into an 8-dimensional space. 
For Rosenbrock tasks, the first linear layer has the same number of nodes (i.e., 18) as the dimensionality of the intermediate output space being embedded. For the composite Langermann function, the first linear layer has 32 nodes. 

\subsubsection{Environmental Modeling Problem}
\paragraph{Problem Setup}
The environmental modeling function is adapted into a high-dimensional problem.
We use the high-dimensional extension of this task used by~\citet{highdimoutputs}. 
This extension allows us to apply JoCo to a version of this function with input dimensionality 15 and output dimensionality 16. 

\paragraph{Encoder Architecture}
For the environmental modeling with JoCo, as with synthetic problems, we use a feed-forward neural network with two linear layers to reduce output spaces. The second layer has 8 nodes, and the first has 16 nodes, matching the intermediate output's dimensionality.

\subsubsection{PDE Optimization Task}
\paragraph{Problem Setup}
The PDE gives two outputs at each grid point, resulting in an intermediate output space with dimensionality $64^2 \cdot 2 = 8192$. We use an input space with $32$ dimensions. Of these, the first four are used to define the four parameters of the PDE while the other $28$ are noise that the optimizer must learn to ignore.
 
\paragraph{Encoder Architecture}
In order to embed the $8192$-dimensional output space with JoCo, we use a simple feed-forward neural net with three linear layers of 256, 128, and 32 nodes, respectively. 
We therefore embed the $8192$-dimensional output space to a $32$-dimensional space. 

\subsubsection{Rover Trajectory Planning}
\paragraph{Problem Setup}
This task is inherently composite in nature as each evaluation allows us to observe both the cost function value and the intermediate output trajectory.
For this task, intermediate outputs are $1000$-dimensional since each trajectory consists of a set of $500$ coordinates in 2D space. 

\paragraph{Encoder Architecture}
In order to embed this $1000$-dimensional output space with JoCo, we use a simple feed forward neural net with three linear layers that have 256, 128, and 32 nodes respectively. We therefore embed the $1000$-dimensional output space to a $32$-dimensional space. 

\subsubsection{Black-Box Adversarial Attack on Large Language Models}
\paragraph{Problem Setup}
For this task, we obtain an embedding for each word in the input prompts using the 125M parameter version of the OPT Embedding model \cite{opt}. The input search space is therefore $3072$-dimensional (4 tokens per prompts * 768-dimensional embedding for each token).
We limit generated text to 100 tokens in length and pad all shorter generated text so that all LLM outputs are 100 tokens long. 
For each prompt evaluated, we ask the LLM to generate three unique outputs and optimize the average utility of the three generated outputs. Optimizing the average utility over three outputs encourages the optimizer to find prompts capable of \emph{consistently} causing the model to generate uncivil text. We take the average $768$-dimensional embedding over the words in the $100$-token text outputs. The resulting intermediate output is $2304$-dimensional (3 generated text outputs * 768-dimensional average embedding per output).  

\paragraph{Encoder Architecture}
In order to embed this $2304$-dimensional output space with JoCo, we use a simple feed forward neural net with three linear layers that have 256, 64 and 32 nodes respectively. We therefore embed the $2304$-dimensional output space to a $32$-dimensional space. 

\subsubsection{Adversarial Attack on Image Generative Models}
\paragraph{Problem Setup}
As in the LLM prompt optimization task, we obtain an embedding for each word in the input prompts using the 125 million parameter version of the OPT Embedding model \cite{opt}. The input search space is therefore $3072$-dimensional (4 tokens per prompts x 768-dimensional embedding for each token).
For each prompt evaluated, we ask the text-to-image model to generate three unique images and optimize the average utility of the three generated images. Optimizing the average utility over three outputs encourages the optimizer to find prompts capable of \emph{consistently} causing the model to generate images of the target class. The resulting intermediate output is therefore $451584$-dimensional (224 x 224 x 3 image dims x 3 total images per prompt).  
Since the intermediate outputs are images, we use a convolutional neural net to embed this output space.

\paragraph{Encoder Architecture}
We use a simple convnet with four 2D convolutional layers, each followed by a 2x2 max pooling layer, and then finally two fully connected linear layers with 64 and 32 nodes respectively. We therefore embed the $451584$-dimensional output space to a $32$-dimensional space. 

\section{Additional Examples for Image Generation Task}
\label{appdx:sec:more-image-gen-examples}
\begin{figure*}[ht]
    \centering
    \includegraphics[width=\linewidth]{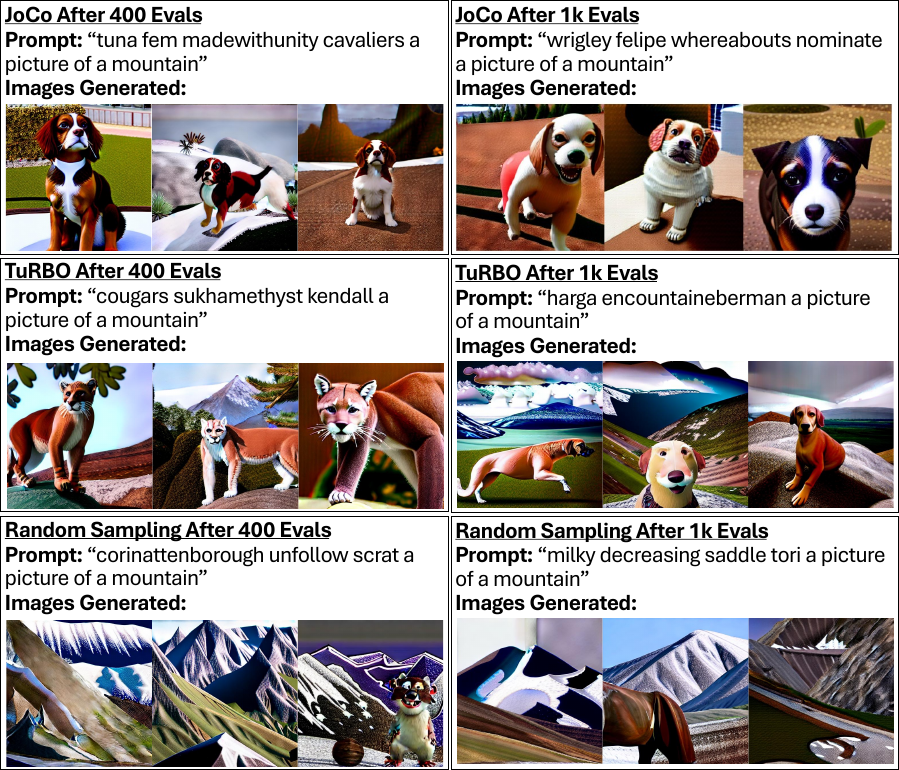 }
    \caption{Examples of the best prompts found by JoCo, TuRBO, and random sampling for the dog image generation task after 400 function evaluations, and after the full budget of 1000 function evaluations. For the dog image generation task, the optimization methods seek to trick a text-to-image model into generating images of dogs despite 1) no individual words related to dogs being present in the prompt and 2) the prompt being pre-pended to the misleading text ``a picture of a mountain". Successful prompts are those that trick the text-to-image model into consistently generating images of dogs. 
    At the full budget of 1000 function evaluations, both JoCo and TuRBO can find prompts that successfully generate images that contain dogs. However, after only 400 function evaluations, only Joco has found a successful prompt, while the best prompt found by TuRBO generates images of cougars rather than dogs. The random sampling baseline is never able to generate pictures with dogs. 
    }
    \label{fig:dog-image-gen-more-examples}
\end{figure*}

In the dog image generation task, the optimizer seeks to find prompts which mislead a text-to-image model to generate images of dogs, despite the absence of individual words related to dogs and despite prompts being pre-pended to the misleading text ``a picture of a mountain''. 
In Figure~\ref{fig:image_gen_examples} b), we provide examples of the best prompts found by JoCo for the dog image generation task after running JoCo for the full budget of 1000 function evaluations. 
In Figure~\ref{fig:dog-image-gen-more-examples}, we additionally include examples of the best prompts found by two baseline methods: TuRBO and random sampling. For all three optimization methods (JoCo, TuRBO, and random sampling), we include examples of the best prompt found by the optimizer after only 400 function evaluations, and after the full budget of 1000 function evaluations. 
As in Figure~\ref{fig:image_gen_examples}, examples in Figure~\ref{fig:dog-image-gen-more-examples} include both the best prompt found by the optimizer and three example images generated when the prompt is given to the text-to-image model. 

At the full budget of 1000 function evaluations, notice that both JoCo and TuRBO can find prompts that successfully generate images that look clearly like dogs. 
However, after only 400 function evaluations, only Joco has found a successful prompt while the best prompt found by TuRBO generates images of cougars rather than dogs. 
This is consistent with results in Figure~\ref{fig:all_tasks} which show that, while TuRBO often eventually converges to a high final reward by the end of the optimization budget, JoCo has significantly better anytime performance, achieving high reward after a much smaller number of function evaluations. 

\end{document}